\documentclass[10pt,twocolumn,letterpaper]{article}

\usepackage{iccv}
\usepackage{times}
\usepackage{epsfig}
\usepackage{graphicx}
\usepackage{amsmath}
\usepackage{amssymb}
\usepackage{subfigure}

\usepackage[breaklinks=true,bookmarks=false]{hyperref}

\iccvfinalcopy 

\def\iccvPaperID{****} 

\ificcvfinal\fi

\begin{document}

\title{Very Long Natural Scenery Image Prediction by Outpainting}

\author{Zongxin Yang$^{1,2}$ \quad Jian Dong$^{3}$ \quad Ping Liu$^2$ \quad Yi Yang$^2$ \quad Shuicheng Yan$^{4}$ \\
$^1$SUSTech-UTS Joint Centre of CIS, Southern University of Science and Technology \\ $^2$ReLER, University of Technology Sydney \quad $^3$ Qihoo 360 \quad $^4$ Yitu Technology\\
\small{zongxin.yang@student.uts.edu.au, dongjian-iri@360.cn, \{ping.liu,yi.yang\}@uts.edu.au, shuicheng.yan@yitu-inc.com}
}

\maketitle
\ificcvfinal\thispagestyle{empty}\fi

\begin{abstract}
Comparing to image inpainting, image outpainting receives less attention due to two challenges in it. The first challenge is how to keep the spatial and content consistency between generated images and original input. The second challenge is how to maintain high quality in generated results, especially for multi-step generations in which generated regions are spatially far away from the initial input. To solve the two problems, we devise some innovative modules, named Skip Horizontal Connection and Recurrent Content Transfer, and integrate them into our designed encoder-decoder structure. By this design, our network can generate highly realistic outpainting prediction effectively and efficiently. Other than that, our method can generate new images with very long sizes while keeping the same style and semantic content as the given input. To test the effectiveness of the proposed architecture, we collect a new scenery dataset with diverse, complicated natural scenes. The experimental results on this dataset have demonstrated the efficacy of our proposed network. The code and dataset are available from \url{https://github.com/z-x-yang/NS-Outpainting}.

\end{abstract}

\section{Introduction}
\begin{figure}[t]
\centering
\includegraphics[width=1.0\linewidth]{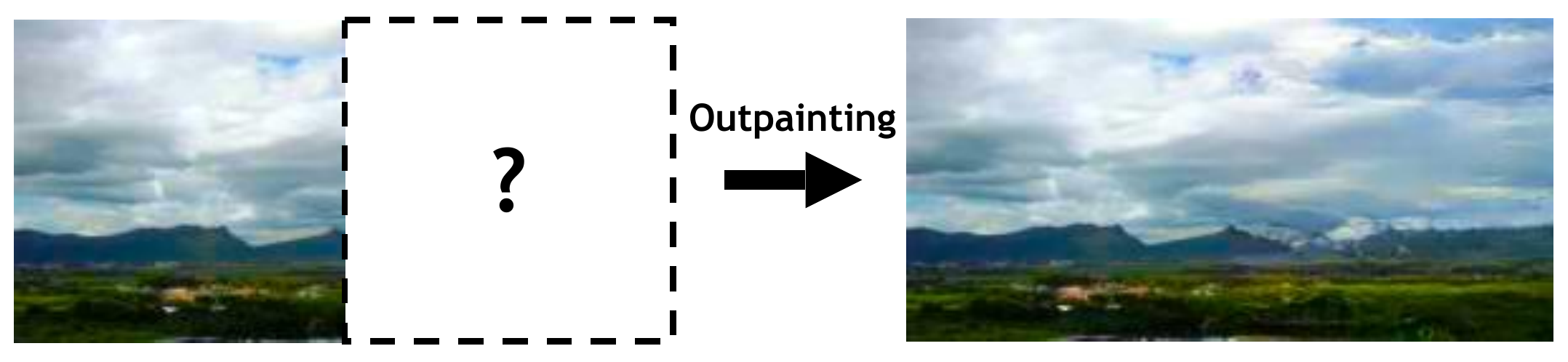}
\caption{Illustration of image outpainting in one step. Given an image as input, image outpainting generates a new image with the same size but outside the original boundary. The spatial configuration and semantic meaning between generated images and the original input must keep consistent.}
\vspace{-3mm}
\label{fig:outpainting}
\end{figure}

Image outpainting, as illustrated in Fig.~\ref{fig:outpainting}, is to generate new contents beyond the original boundaries for a given image. The generated image should be consistent with the given input, both on spatial configuration and semantic content. Although image outpainting can be used in various applications, the solutions with promising results are still in shortage due to the difficulties of this problem. 

The difficulties for image outpainting exist in two aspects. First, it is not easy to keep the generated image consistent with the given input in terms of the spatial configuration and semantic content. Previous works, e.g., ~\cite{biggerpic} needs local warping to make sure there is no sudden change between the input image and the generated region, especially around the boundaries of the two images. Second, it is hard to make the generated image look realistic since it has less contextual information comparing with image inpainting~\cite{traditional_inpainting1,traditional_inpainting2}. 



\begin{figure}[t]
\centering
\includegraphics[width=1.0\linewidth]{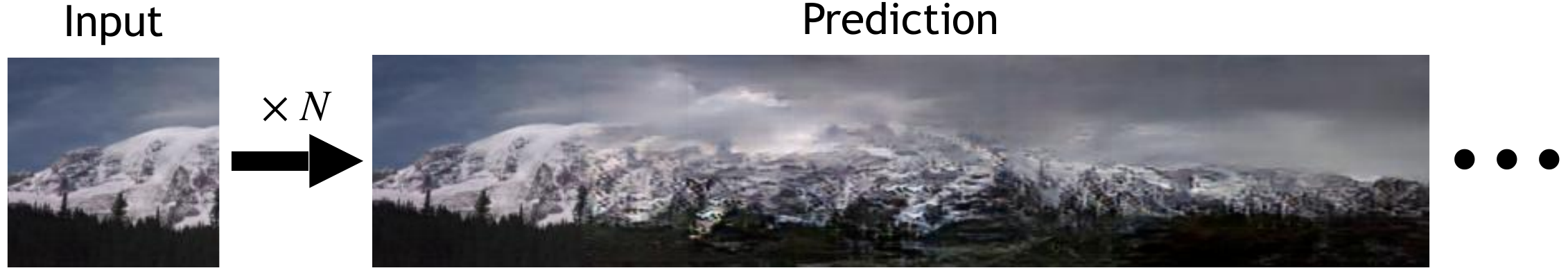}

\caption{Illustration of image outpainting for natural scenery images horizontally in multi-steps.}
\vspace{-5mm}
\label{fig:outpainting_ms}
\end{figure}

For solving image outpainting problems, a few preliminary works were published~\cite{kopf2012quality, sivic2008creating, zhang2013framebreak, biggerpic}. However,  none of those works~\cite{kopf2012quality, sivic2008creating, zhang2013framebreak, biggerpic} utilize ConvNets. Those works attempt to  ``search" image patch(es) from given candidates,  concatenate the best match(es) with the original input spatially.  Those works have their limitations: (1) they need handcrafted features to summarize the image; (2) they need image processing techniques, for example, local warping~\cite{biggerpic}, to make sure there is no sudden visual change between input and generated images; (3) the final performance is heavily dependent on the size of the candidate pool.

Inspired by the success of deep networks on inpainting problems~\cite{context_encoders}, we draw on a similar encoder-decoder structure with a global and a local adversarial loss, to solve image outpainting. In our architecture, the encoder is to compress the given input into a compact feature representation, and the decoder generates a new image based on the compact feature representation. More than that, to solve the two challenging problems in image outpainting, we make several innovative improvements in our architecture.


To make the generated images spatial and semantic consistent with original input, it is necessary to take full advantages of the information from the encoder and fuse it into the decoder.  For this purpose, we design a \textbf{Skip Horizontal Connection (SHC)} to connect encoder and decoder at each same level. By this way,  the decoder can generate a prediction with strong regards to the input. Our experimental results prove that the proposed SHC can improve the smoothness and reality of the generated image.

Moreover, we propose \textbf{Recurrent Content Transfer (RCT)}, to transfer the sequence from the encoder to the decoder to generate new contents.  Compared to channel-wise full connection strategy in the previous work~\cite{context_encoders},  RCT can facilitate our network to handle the spatial relationship in the horizontal direction more effectively. Besides, by adjusting the length of the prediction feature, RCT assists our architecture in controlling the prediction size conveniently, which is hard if utilizing full connection.



By integrating the proposed \textit{SHC} and \textit{RCT} into our designed encoder-decoder architecture, our method can successfully generate images with~\textit{extra length} outside the boundary of the given image. As shown in Figure.~\ref{fig:outpainting_ms}, it is a recursive process since the generation from the last step is utilized as the input for the current step, which, theoretically, can generate smooth, and realistic images with a very long size. Those generated images, although spatially far away from the given input and thus receiving little contextual information from it, still keep high qualities. 

To demonstrate the effectiveness of our method, we collect a new scenery dataset with $6,000$ images, which consists of diverse, complicated natural scenes, including mountain with or without snow, valley, seaside, riverbank, starry sky, etc. We conduct a series of experiments on this dataset and not surprisingly beat all competitors~\cite{pix2pix2017, glc, generative_inpainting}. 

 \textbf{Contributions.} Our contributions are summarized in the following aspects:

(1)~we design a novel encoder-decoder framework to handle image outpainting, which is rarely discussed before;

(2)~we propose Skip Horizontal Connection and Recurrent Content Transfer, and integrate them into our designed architecture, which not only significantly improves the consistency on spatial configuration and semantic content, but also enables our architecture with an excellent ability for long-term prediction;


(3)~we collect a new outpainting dataset, which has $6,000$ images containing complex natural scenes. We validate the effectiveness of our proposed network on this dataset.
\section{Related Work}

In this section, we briefly review the previous works relating to this paper in five sub-fields: Convolutional Neural Networks, Generative Adversarial Networks, Image Inpainting, Image Outpainting, and Image-to-Image Translation.

\textbf{Convolutional Neural Networks (ConvNets)} VGGNets\cite{vggnet} and Inception models \cite{inception} demonstrate the benefits of deep network. To train deeper networks, Highway networks \cite{highway}
employ a gating mechanism to regulate shortcut connections. ResNet \cite{ResNet} simplifies the shortcut connection and shows the effectiveness of learning deeper networks through the use of identity-based skip connections. Due to the complexity of our task, we employ a group of "bottleneck" ResBlocks \cite{ResNet} to build our network and utilize residual connections in Skip Horizontal Connection to improve the smoothness of the generated results.

\textbf{Generative Adversarial Networks (GANs)} GAN~\cite{gan} has achieved success in various problems, including image generation~\cite{image_generation1,image_generation2}, image inpainting~\cite{context_encoders}, future prediction~\cite{future_prediction}, and style translation~\cite{cycle_gan}. The key to the success of GANs is the introduction of the adversarial loss, which forces the generator to captures the true data distribution. To improve the training of GAN,  variants of GANs have been derived.  For example,  WGAN-GP~\cite{wgan_gp} introduces a gradient penalty and achieves more stable training.  And thus we utilize WGAN-GP in this work due to its advantages.

\begin{figure*}[ht]
\centering

\subfigure[Overview]{
\begin{minipage}[b]{0.51\linewidth}
\includegraphics[width=1\textwidth]{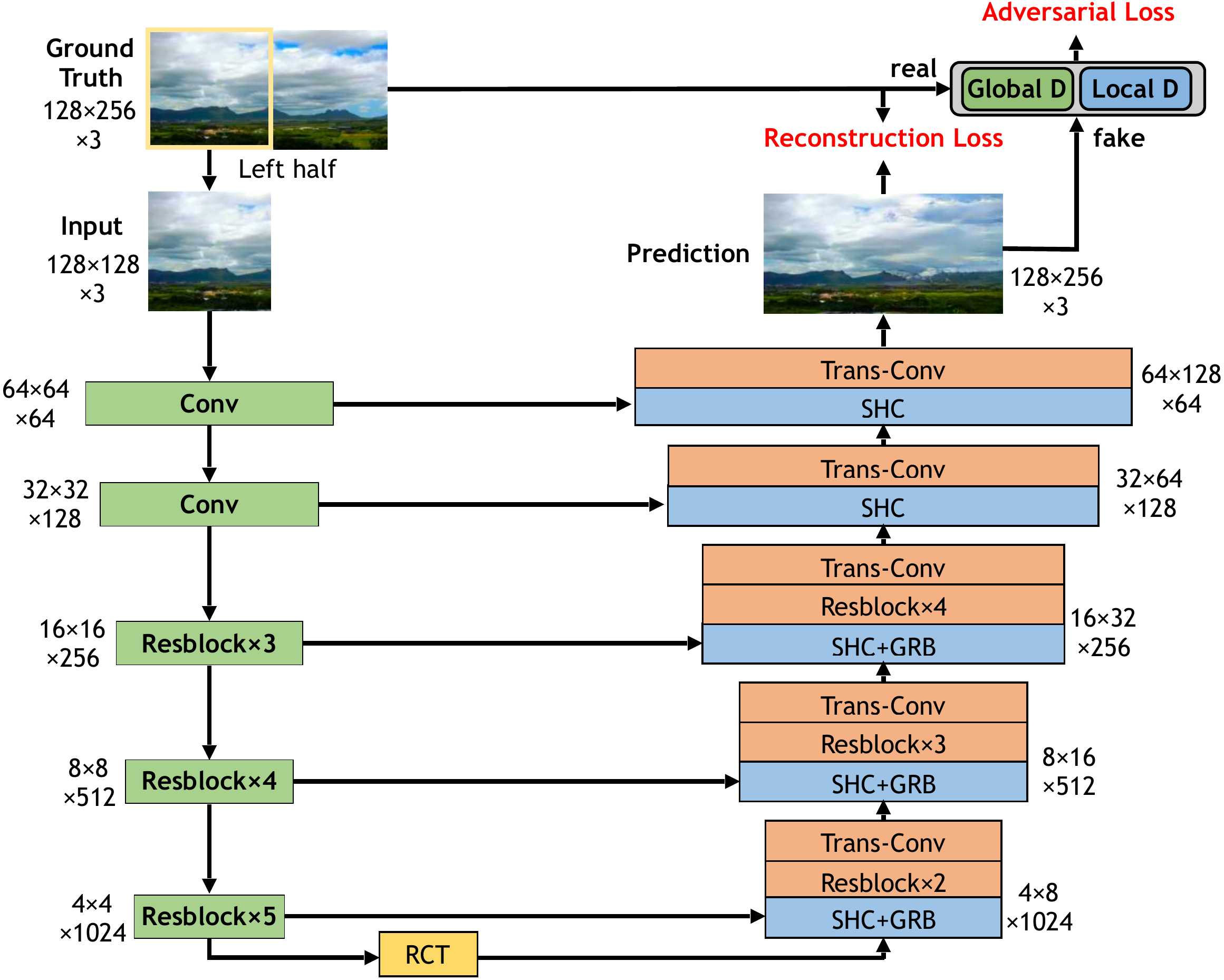}
\end{minipage}
\label{fig:3.a}}
\subfigure[Multi-Step Generation]{
\begin{minipage}[b]{0.36\linewidth}
\includegraphics[width=1\textwidth]{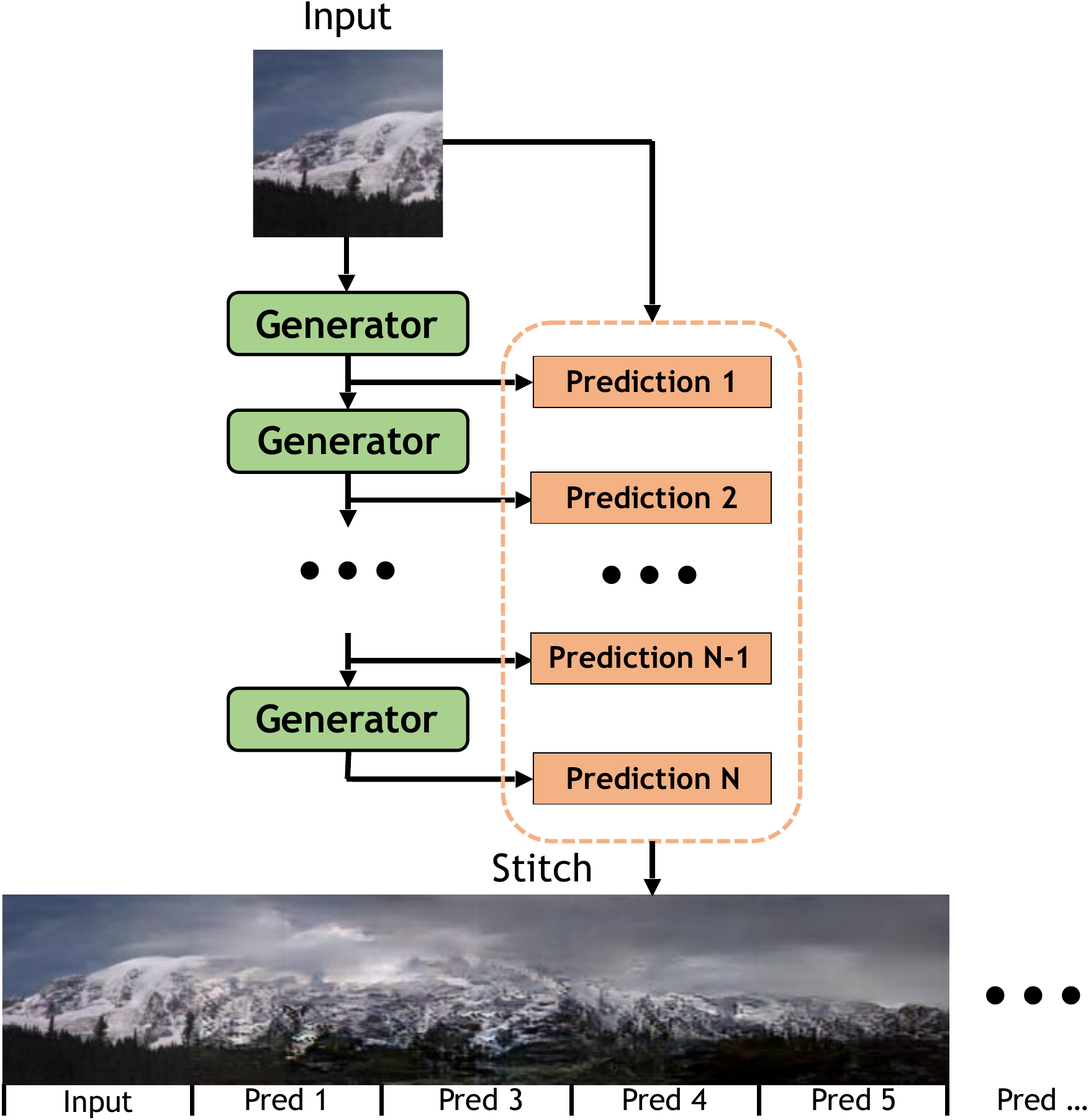}
\end{minipage}
\label{fig:3.b}}

\caption{(a) The overall architecture consists of a generator and a discriminator. The generator exploits an encoder-decoder pipeline. We propose Recurrent Content Transfer (RCT) to link the encoder and decoder. Meanwhile, We deploy 
Skip Horizontal Connection (SHC) to connect the encoder and decoder at each symmetrical level. Moreover, after the first three SHC layers, we deploy Global Residual Blocks (GRB), which has a large receptive field, to further strengthen the connection between the predicted and original region. (b) We can generate an image with very long sizes by iterating the generator.}
\label{fig:3}
\end{figure*}

\textbf{Image Inpainting} The classical image inpainting \cite{traditional_inpainting1,traditional_inpainting2} approaches utilize local non-semantic methods to predict the missing region. However, when the missing region size becomes huge, or the context grows complex,  the quality of the final results deteriorates~\cite{context_encoders,inpainting2,glc, generative_inpainting}. Compared to image inpainting, image outpainting is more challenging. To the best of our knowledge, there is NO other peer-reviewed published work utilizing ConvNets for image outpainting before our work.

\textbf{Image Outpainting} There are a few preliminary published works~\cite{kopf2012quality, sivic2008creating, zhang2013framebreak, wang2014biggerpicture} for image outpainting problems, but none of them utilized ConvNets. Those works employed image matching strategies to ``search" image patch(es) from the input image or an image library, and treat the patch(es) as prediction regions. If the search fails, the final ``prediction" result will be inconsistent with the given context.  Unlike those previous work~\cite{kopf2012quality, sivic2008creating, zhang2013framebreak, wang2014biggerpicture}, our approach does not need any image matching strategy but depends on our carefully designed deep network.

\textbf{Image-to-Image Translation}
With the development of ConvNets,  recent approaches~\cite{pix2pix2017,gan,sketch,cycle_gan} for image-to-image translation design deep networks for learning a parametric translation function.  After ``Pix2Pix"~\cite{pix2pix2017} framework, which use a conditional adversarial network \cite{gan} to learn a mapping from input to output images, similar ideas have been applied to related tasks, such as translating sketches to photographs \cite{sketch}, style translation \cite{cycle_gan,dong2018san}, etc. Although image outpainting is similar to the image-to-image translation task, there is a significant difference between them: for image-to-image translation, the input and output keep the same semantic content but change details or styles; for our work, the style is shared between the input and output,  the semantic contents are different but keep consistent.

\section{Methodology}
We first provide an overview of the overall architecture, which is shown in Fig.~\ref{fig:3}, then provide details on each component.

\subsection{Encoder-Decoder Architecture}

\begin{table}[h] 
  \centering  
    \begin{tabular}{c|c|c}  
    \hline
    \textbf{layer} & \textbf{output size} & \textbf{parameters}\\ 
    \hline
      Conv & 64$\times$64$\times$64 & 4$\times$4, stride=2 \\
    \hline
      Conv & 32$\times$32$\times$128 & 4$\times$4, stride=2 \\
    \hline
      ResBlock$\times$3 & 16$\times$16$\times$256 & stride of first block=2 \\
    \hline
      ResBlock$\times$4 & 8$\times$8$\times$512 & stride of first block=2 \\
    \hline
      ResBlock$\times$5 & 4$\times$4$\times$1024 & stride of first block=2 \\
    \hline
      RCT & 4$\times$4$\times$1024 & None \\
    \hline
      SHC+GRB & 4$\times$8$\times$1024 & dilated rate=1 \\
    \hline
      ResBlock$\times$2 & 4$\times$8$\times$1024 & None \\
    \hline
      Trans-Conv & 8$\times$16$\times$512 & 4$\times$4, stride=2 \\
    \hline
      SHC+GRB & 8$\times$16$\times$512 & dilated rate=2 \\
    \hline
      ResBlock$\times$3 & 8$\times$16$\times$512 & None \\
    \hline
      Trans-Conv & 16$\times$32$\times$256 & 4$\times$4, stride=2 \\
    \hline
      SHC+GRB & 16$\times$32$\times$256 & dilated rate=4 \\
    \hline
      ResBlock$\times$4 & 16$\times$32$\times$256 & None \\
    \hline
      Trans-Conv & 32$\times$64$\times$128 & 4$\times$4, stride=2 \\
    \hline
      SHC & 32$\times$64$\times$128 & None \\
    \hline
      Trans-Conv & 64$\times$128$\times$64 & 4$\times$4, stride=2 \\
    \hline
      SHC & 64$\times$128$\times$64 & None \\
    \hline
      Trans-Conv & 128$\times$256$\times$3 & 4$\times$4, stride=2 \\
    \hline
    \end{tabular}
  
  \caption{The specific parameters of generator. Trans-Conv is transposed convolution.} 
  \label{tab:1}
  \label{table1} 
\end{table}

We design an encoder-decoder architecture for image outpainting. Our encoder takes an input image and extracts its latent feature representation; the decoder takes this latent representation to generate a new image with the same size, which has consistent content and the same style.

\textbf{Encoder} Our encoder is derived from the \textit{ResNet-50} \cite{ResNet}. The difference is that we replace \textit{max pooling} layers with convolutional layers, and remove layers after \textit{conv4\_5}. Given an input image \textit{I} of size 128$\times$128, the encoder will compute a latent representation with the dimension of 4$\times$4$\times$1024. 

\begin{figure*}[ht]
\centering

\subfigure[Ground Truth]{
\begin{minipage}[b]{0.18\linewidth}
\includegraphics[width=1\linewidth]{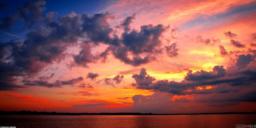}
\includegraphics[width=1\linewidth]{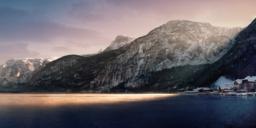}
\end{minipage}
\label{fig:7.a}}
\subfigure[FC]{
\begin{minipage}[b]{0.18\linewidth}
\includegraphics[width=1\linewidth]{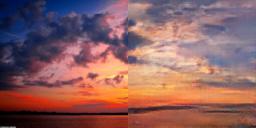}
\includegraphics[width=1\linewidth]{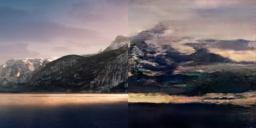}
\end{minipage}
\label{fig:7.b}}
\subfigure[FC+SHC]{
\begin{minipage}[b]{0.18\linewidth}
\includegraphics[width=1\linewidth]{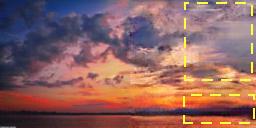}
\includegraphics[width=1\linewidth]{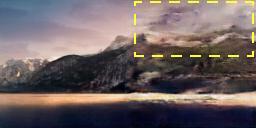}
\end{minipage}
\label{fig:7.c}}
\subfigure[RCT+SHC(Ours)]{
\begin{minipage}[b]{0.18\linewidth}
\includegraphics[width=1\linewidth]{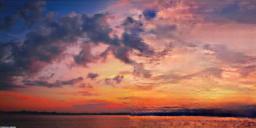}
\includegraphics[width=1\linewidth]{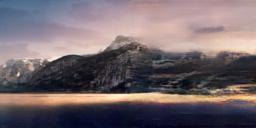}
\end{minipage}
\label{fig:7.d}}

\caption{The qualitative results on our collected scenery dataset. The method of (b) uses a fully-connected (FC) layer to connect the encoder and decoder, with which an obvious un-smoothness on the boundary between original and predicted regions. And the method of (c) deploys SHC layers to mitigate the un-smoothness, but there is still a problem that the generated image is easily getting blurred when the prediction region is far away from the input region. We use yellow boxes to highlight the blurred areas in predicted areas. Finally, the method of (d), which replaces the FC layer with RCT, overcomes the problem in (c) and makes the details in the prediction more delicate.}
\label{fig:7}
\end{figure*}

As pointed out in~\cite{context_encoders}, it is difficult only to utilize convolutional layers to propagate information from input image feature maps to predicted feature maps. The reason is that there is no one-to-one correspondence between them under this circumstance. In Context Encoders~\cite{context_encoders}, this information propagation is handled by \textit{channel-wise fully-connected} (FC) layers. One of the limitations in FC layers is they can only handle features of fixed sizes. In our practice, this limitation will make predicted results deteriorate when the input size is large (Fig.~\ref{fig:7.b}). More than that, as illustrated by~\cite{context_encoders}, FC layers occupy a huge amount of parameters, which makes the training inefficient or impractical. To deal with those problems, we propose a Recurrent Content Transfer (RCT) layer for information propagation in our network.

\begin{figure}[t]
\centering
\includegraphics[width=0.8\linewidth]{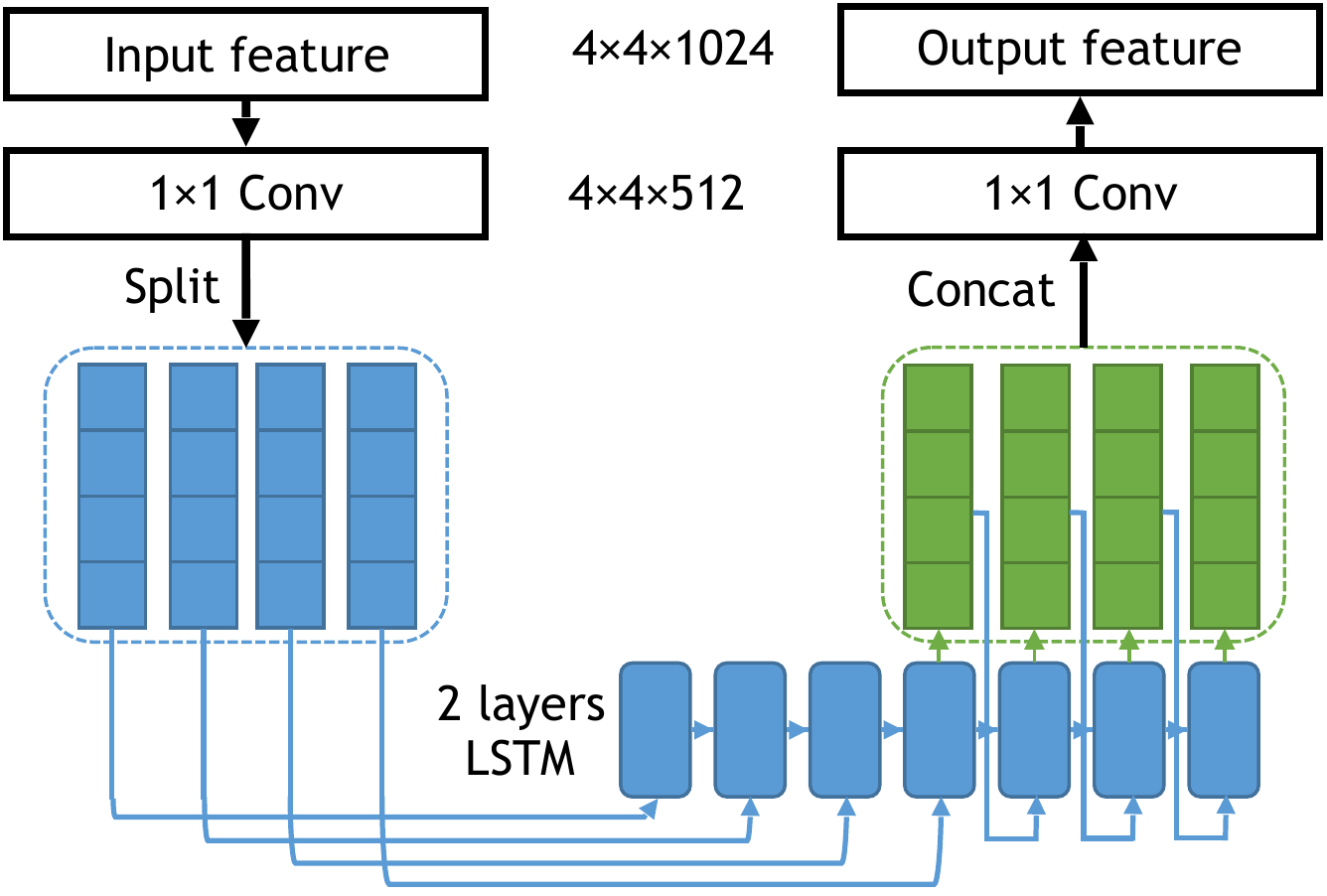}

\caption{The illustration of Recurrent Content Transfer (RCT). 1$\times$1 convolutional layers are utilized to adjust the channel dimension of input and output of RCT. RCT splits the feature representation of input to a sequence in the horizontal direction, and uses two \textit{LSTM} layers to transfer this sequence to a predicted sequence. The size of the prediction region can be adjusted by setting the length of the prediction sequence in 1-step prediction, which is set to 4 to achieve a satisfactory result in our practice.}
\label{fig:4}
\end{figure}

\textbf{Recurrent Content Transfer} RCT, which is shown in Figure.~\ref{fig:4}, is designed for efficient information propagation between feature sequences from input regions and prediction regions respectively. Specifically, RCT splits the feature maps from the input region to a sequence in the horizontal dimension, and then uses two \textit{LSTM} \cite{lstm} layers to transfer this sequence to a new sequence corresponding to the prediction region. After that, the new sequence is concatenated and reshaped into predicted feature maps. 1$\times$1 convolutional layers are utilized to adjust the channel dimensions of input and output in RCT.  
Given input feature maps with a size of 4$\times$4$\times$1024, RCT outputs feature maps with the same dimension. 

Benefit from the recurrent structure in RCT, we can control the size of the prediction region by setting the length of the prediction sequence in 1-step prediction. And by iterating the model, we can generate images with high-quality and very long range (Fig.~\ref{fig:9},~\ref{fig:10}). 

\begin{figure*}[ht]
\centering

\subfigure[SHC]{
\begin{minipage}[b]{0.385\linewidth}
\includegraphics[width=1\textwidth]{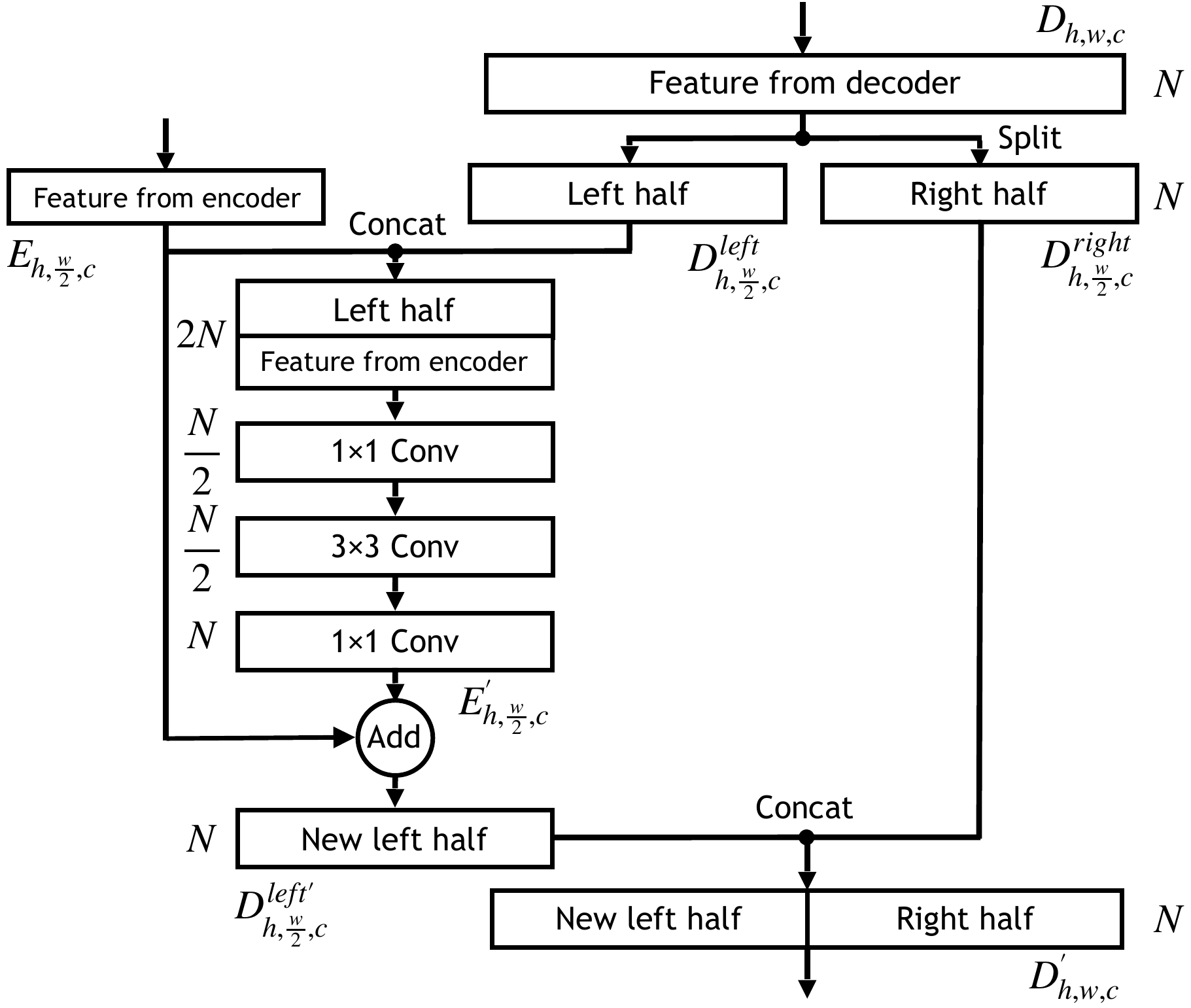}
\end{minipage}
\label{fig:5.a}}
\subfigure[GRB]{
\begin{minipage}[b]{0.385\linewidth}
\includegraphics[width=1\textwidth]{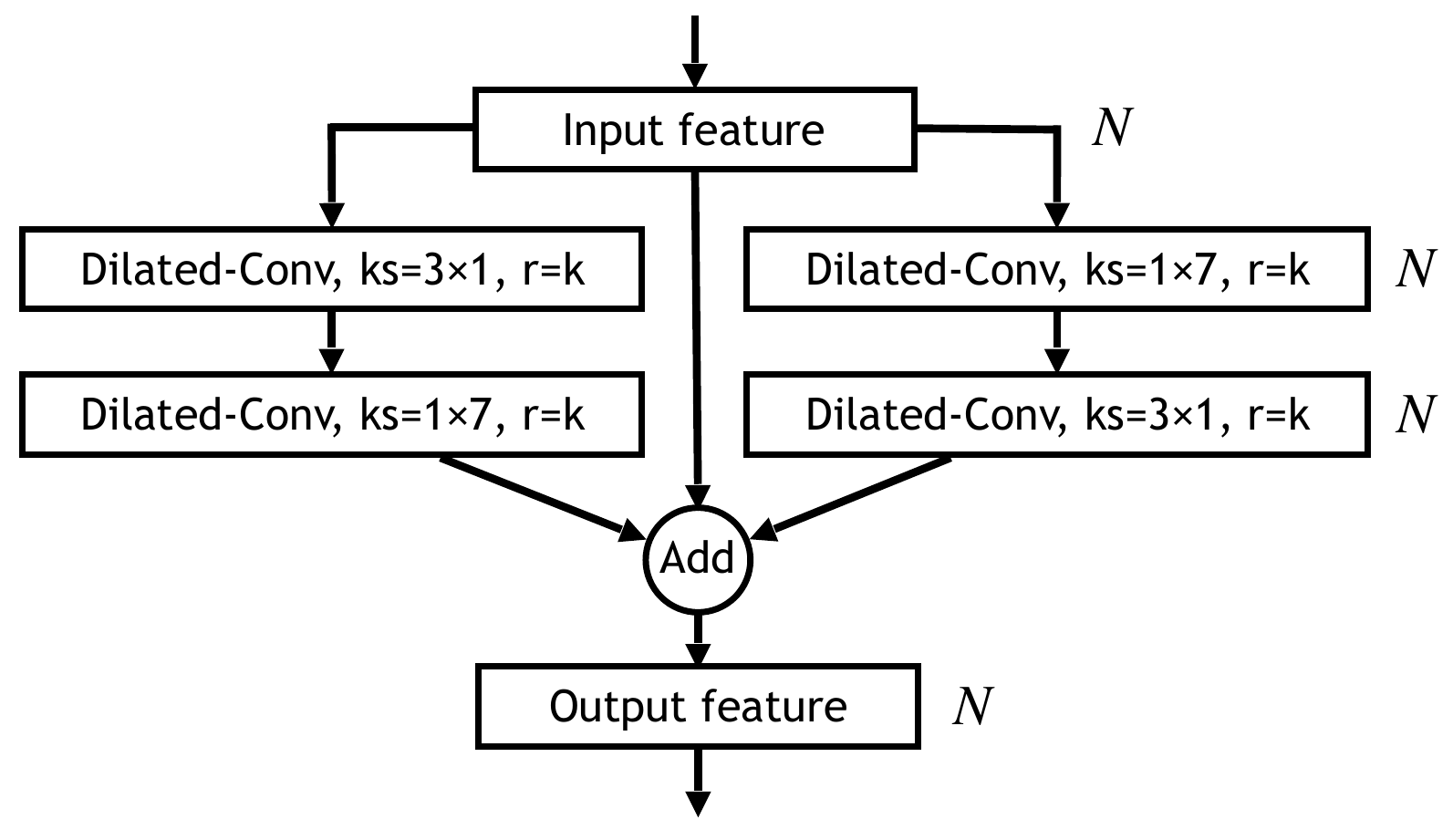}
\end{minipage}
\label{fig:5.b}}

\caption{The details of Skip Horizontal Connection (a) and Global Residual Block (b). $N$ is the channel number, $ks$ is the kernel size, and $r$ is the dilation rate of \textit{dilated-convolutional} layers. In (b), we set a bigger size of receptive field on horizontal dimension (1$\times$7) to strengthen the connection between the input and predicted region.}
\label{fig:5}
\end{figure*}









\begin{figure*}[t!]
\center
\subfigure[No SHC]{
\label{fig:6.a}
\includegraphics[width=0.175\linewidth]{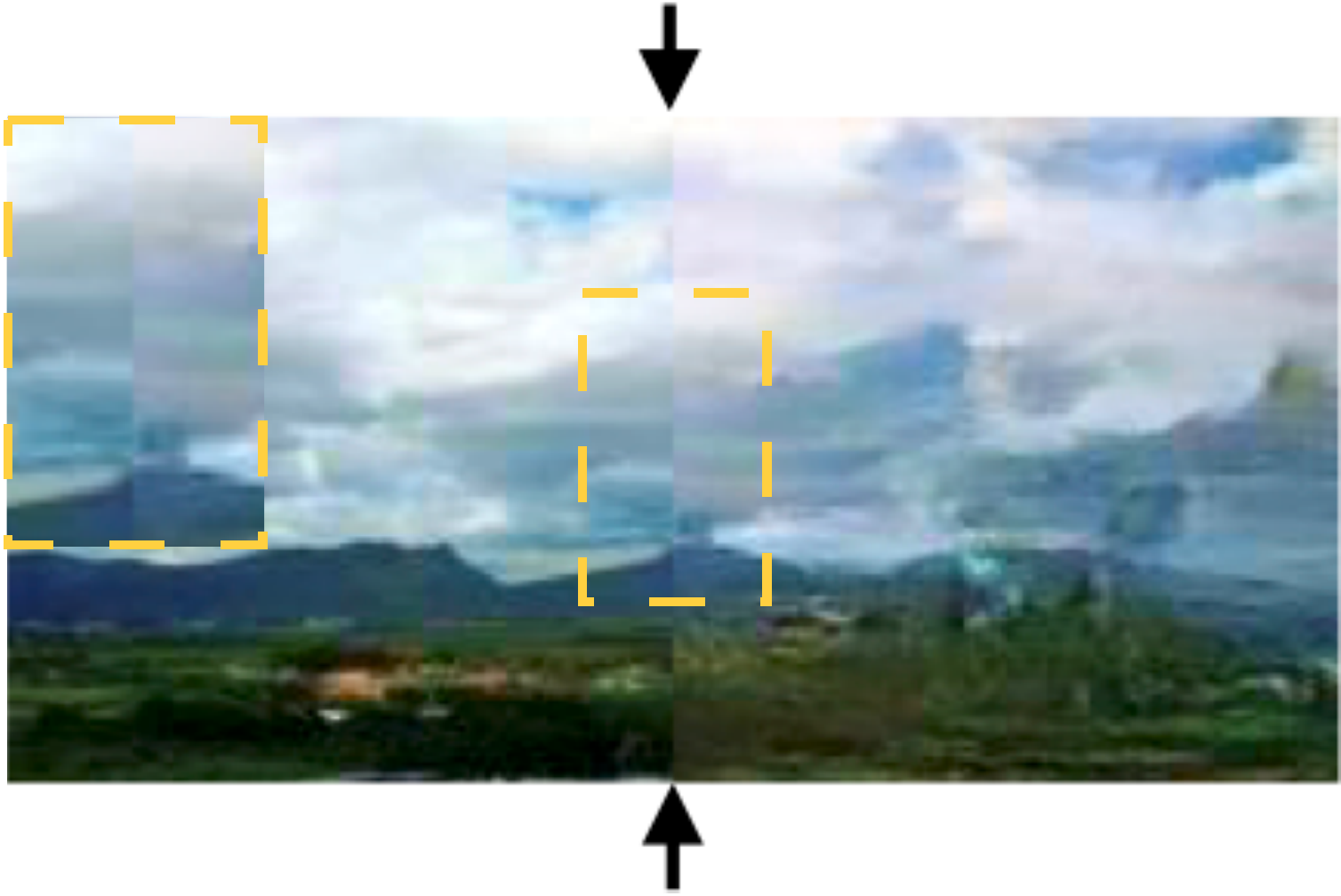}
}
\subfigure[One SHC layer]{
\label{fig:6.b}
\includegraphics[width=0.175\linewidth]{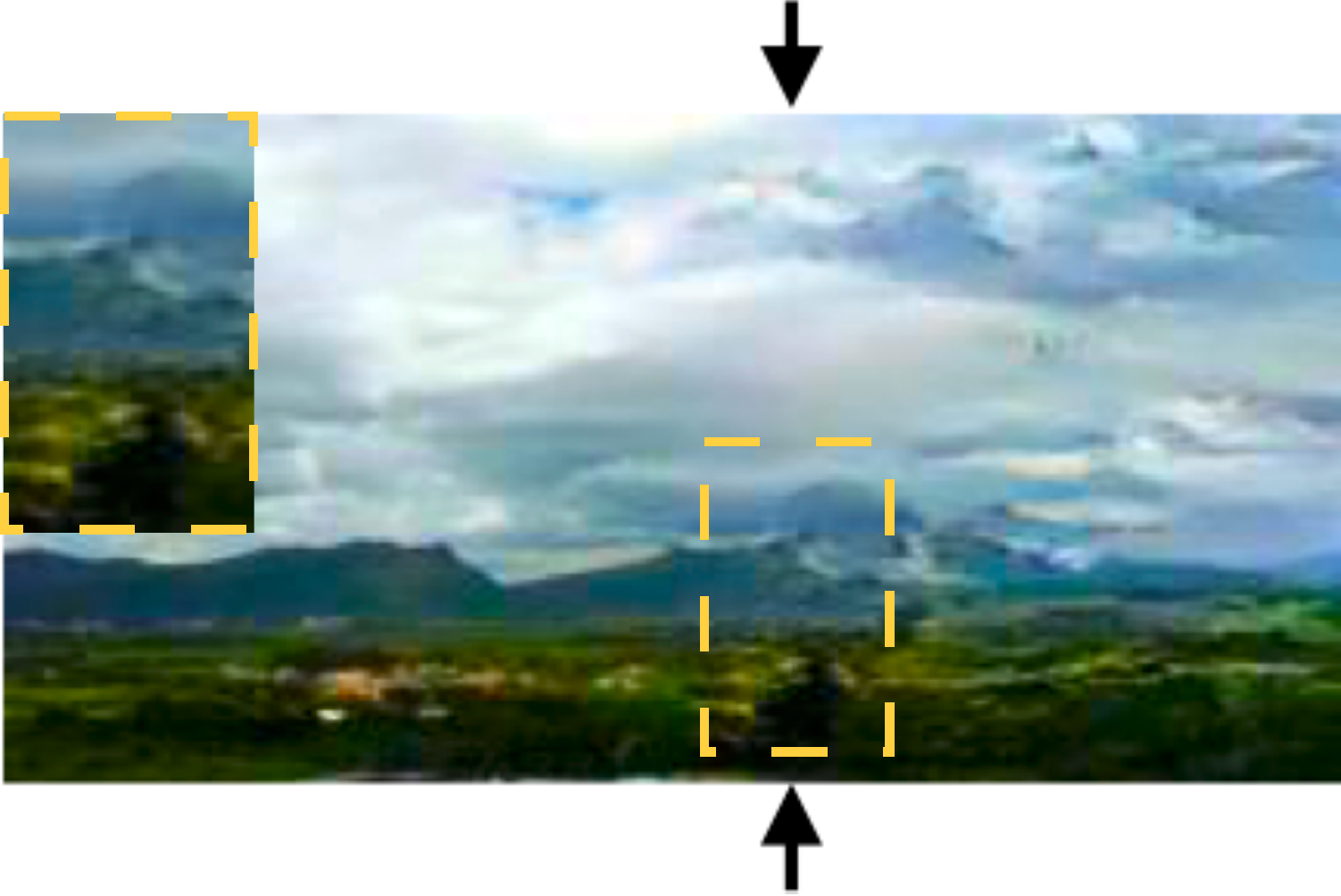}
}
\subfigure[Full SHC]{
\label{fig:6.c}
\includegraphics[width=0.175\textwidth]{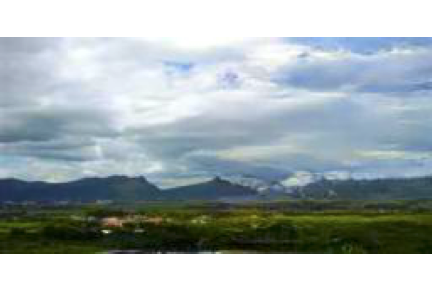}
}
\subfigure[Groundtruth]{
\label{fig:6.d}
\includegraphics[width=0.175\linewidth]{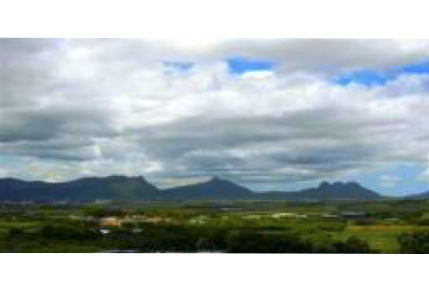}
}

\caption{
(a): When we don't use any SHC layers, there is an obvious boundary between the input region and predicted region, which means an inconsistency during the generation process. (b) When we utilize one SHC layer with GRB in the middle of decoder, the boundary line starts to fade away. (c) After we deploy more SHC layers, there is no obvious boundary.
}
\label{fig:norm_dist}
\end{figure*}

\textbf{Decoder} Decoder takes 4$\times$4$\times$1024 dimensional features, which are encoded from a 128$\times$128 image (\textit{I}), to generate an image of size 128$\times$256. The left half of the generated image is the same as the input image \textit{I}; the right half is predicted by our architecture. Similar to the most recent methods, we use five \textit{transposed-convolutional} layers
~\cite{transposed} in the decoder to expand the spatial size and reduce the channel number. However, unlike the previous work~\cite{context_encoders}, before each \textit{transposed-convolutional} layer, we propose to use our designed \textbf{Skip Horizontal Connection (SHC)} to fuse the feature from the encoder into the decoder. 


\textbf{Skip Horizontal Connection} Inspired by U-Net~\cite{unet}, we propose SHC, which is shown in Fig.~\ref{fig:5.a}, to share information from the encoder to the decoder at the same level. The difference between SHC and U-Net \cite{unet} is that the spacial size of the encoder feature is different from the decoder in SHC. SHC focuses on the left half of the decoder feature which corresponds to the original input region. 

As illustrated in Fig.~\ref{fig:5.a}, given a feature $D_{h,w,c}$ from decoder and a feature $E_{h,\frac{w}{2},c}$ from encoder, SHC computes a new feature $D_{h,w,c}^{'}$. The procedures are as follows: First, we concatenate the left half of $D_{h,w,c}$, denoted as $D_{h,\frac{w}{2},c}^{left}$, with $E_{h,\frac{w}{2},c}$ on the channel dimension; then, we pass this concatenated feature through three convolutional layers, which have kernels of 1$\times$1, 3$\times$3 and 1$\times$1 size respectively, to get to a feature representation denoted as $E_{h,\frac{w}{2},c}^{'}$. To make the training more stable, we introduce a residual connection to make a element-wise addition between $E_{h,\frac{w}{2},c}^{'}$ and $E_{h,\frac{w}{2},c}$. We denote the addition result as $D_{h,\frac{w}{2},c}^{left'}$. We use $D_{h,\frac{w}{2},c}^{left'}$ to replace the left half of the input feature for SHC, $D_{h,w,c}$, to get the final output for SHC, denoted as $D_{h,w,c}^{'}$.\footnote{Specially, the SHC before first \textit{transposed-convolutional} layer is different from above. In this layer, we just concatenate the input of RCT to the left of predicted feature map on width dimension, because the predicted feature doesn't include any information from the input region to compute.}



Besides, to keep a balance between the insufficient context due to small kernel sizes and the high computation cost introduced by large kernel dimensions, we propose to combine the advantage of Residual Block \cite{ResNet} and \textit{Inception} into a novel block: \textit{Global Residual Block (GRB)}, which is shown in Figure.~\ref{fig:5.b}.

In GRB, a combination of 1$\times$n and n$\times$1 convolutional layers replace n$\times$n convolutional layers, the residual connection is introduced to connect the input to output, and \textit{dilated-convolutional} layers~\cite{dilated} is utilized to ``support exponential expansion of the receptive field without loss of resolution or coverage". To strengthen the connection between the original and predicted region aligned on the horizontal direction, we set a bigger receptive field on the horizontal dimension in GRB. \footnote{We only deploy GRB after first three SHC layers, because we found it fails to achieve good performance when setting GRB too close to the output layer. After GRB, we deploy some ResBlocks to compensate for the performance loss caused by \textit{Inception} architecture and \textit{Dilated convolutions}.}

\subsection{Loss Function}
Our loss function consists of two parts: a masked reconstruction loss and an adversarial loss. The reconstruction loss is responsible for capturing the overall structure of the predicted region and logical coherence with regards to the input image, which focuses on low-order information. The adversarial loss \cite{gan,wgan,wgan_gp} makes prediction look more real, which is due to high-order information capturing.

\begin{figure*}[!t]
\centering
\includegraphics[width=0.77\linewidth]{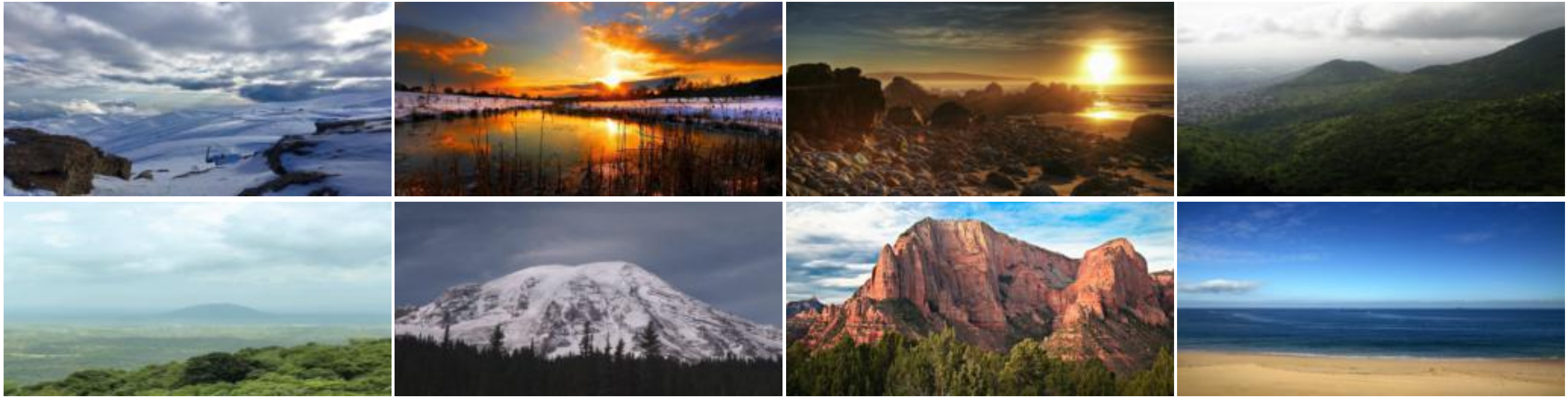}

\caption{Examples of our scenery dataset. This scenery dataset consists of diverse, complicated natural scenes, including mountain with or without snow, valley, seaside, riverbank, starry sky, etc.}
\label{fig:8}
\end{figure*}

\textbf{Masked Reconstruction Loss}
We use a L2 distance between ground truth image $x$ and predicted image $\tilde{x}$ as our reconstruction loss, denoted as $\mathcal{L}_{rec}(x)$,
\begin{equation}
    \mathcal{L}_{rec}(x)=M\odot\parallel x-\tilde{x}\parallel^{2}_{2},
\end{equation}
where $M$ is a mask used to reduce the weights of L2 along the prediction direction. Masked reconstruction loss is prevalent in generative image inpainting task~\cite{context_encoders,glc,generative_inpainting}, because less relation is between ground truth and prediction when far away from the border. Different from other mask methods, we use a $cos$ function to decay the weight to zero. In the predicted region, let $d$ be the distance to the border between origin and predicted region and $W_p$ be the width of prediction in 1-step, we have:
\begin{equation}
M(d)=\frac{1+cos(\frac{d\pi}{W_p})}{2}.
\end{equation}
The L2 loss can minimize the mean pixel-wise error, which makes the generator to produce a rough outline of the predicted region but results in a blurry averaged image \cite{context_encoders}. To alleviate this blurry problem, we add an adversarial loss to capture high-frequency details. 

\textbf{Global and Local Adversarial Loss} 
Following the same strategy utilized in~\cite{generative_inpainting}, we deploy one global adversarial loss and one local adversarial loss, to make the generated images indistinguishable from the real input image. We choose a modified Wasserstein GANs \cite{wgan_gp} for our global and local adversarial loss due to its advantages, the only difference between the global and the local adversarial loss is their input.

Specifically, by enforcing a soft version of the constraint with a penalty on the gradient norm for random samples $\tilde{x}\sim\mathbb{P}_{\tilde{x}}$, the final objective in~\cite{wgan_gp} becomes:
\begin{multline}
    \max\limits_{G}\min\limits_{D}\mathop{\mathbb{E}}\limits_{\tilde{x}\sim\mathbb{P}_{g}}[D(\tilde{x})]-\mathop{\mathbb{E}}\limits_{x\sim\mathbb{P}_{r}}[D(x)] \\
    +\lambda_{gp}\mathop{\mathbb{E}}\limits_{\tilde{x}\sim\mathbb{P}_{\tilde{x}}}[(\parallel \nabla_{\tilde{x}}D(\tilde{x})\parallel_{2}-1)^{2}].
\end{multline}

Hence the adversarial loss for the discriminator, $\mathcal{L}_{dis}$, is
\begin{multline}
    \mathcal{L}_{dis}=\min\limits_{D}\mathop{\mathbb{E}}\limits_{\tilde{x}\sim\mathbb{P}_{g}}[D(\tilde{x})]-\mathop{\mathbb{E}}\limits_{x\sim\mathbb{P}_{r}}[D(x)] \\
    +\lambda_{gp}\mathop{\mathbb{E}}\limits_{\tilde{x}\sim\mathbb{P}_{\tilde{x}}}[(\parallel \nabla_{\tilde{x}}D(\tilde{x})\parallel_{2}-1)^{2}].
\end{multline}

And the adversarial loss for the generator, $\mathcal{L}_{gen}$, is
\begin{equation}
    \mathcal{L}_{gen}=\min\limits_{G}-\mathop{\mathbb{E}}\limits_{\tilde{x}\sim\mathbb{P}_{g}}[D(\tilde{x})]
\end{equation}

In the global adversarial loss, $\mathcal{L}_{dis}^{global}$ and $\mathcal{L}_{gen}^{global}$, the $x$ and $\tilde{x}$ are the ground truth images and the entire output (including original input on the left, and the predicted region on the right). In the local adversarial loss, $\mathcal{L}_{dis}^{local}$ and $\mathcal{L}_{gen}^{local}$, the $x$ and $\tilde{x}$ are the right half of ground truth images and the right half of entire output (the predicted region).

In a summary, the entire loss for global and local discriminators, $\mathcal{L}_{D}$, is
\begin{equation}
    \mathcal{L}_{D} = \beta\mathcal{L}_{gen}^{global} + (1-\beta)\mathcal{L}_{gen}^{local}.\label{con:6}
\end{equation}
And the entire loss for the generator, $\mathcal{L}_{G}$, is
\begin{equation}
    \mathcal{L}_{G}=\lambda_{rec}\mathcal{L}_{rec}+\lambda_{adv}\mathcal{L}_{D}\label{con:7}.
\end{equation}



In our experiments, we set $\lambda_{gp}=10$, $\beta=0.9$, $\lambda_{adv}=0.002$, and $\lambda_{rec}=0.998$.



\subsection{Implementation Details}
In our architecture, we use ReLU as the activation function in the decoder module, Leaky-ReLU as the activation function in other modules. We choose Instance normalization~\cite{instance} instead of Batch normalization~\cite{batch} before these activation functions empirically.

\section{Experiments}
We prepare a new scenery dataset consisting of diverse, complicated natural scenes, including mountain with or without snow, valley, seaside, riverbank, starry sky, etc. There are about $5,000$ images in the training set and $1,000$ images in the testing set. Part of the dataset (about $3,000$) comes from SUN dataset~\cite{sun}, and we collect others on the internet. Fig.~\ref{fig:8} shows some examples.  We conduct a series of comparative experiments to test our model on 1-step prediction\footnote{In our experiment, we do natural scenery image outpainting only on horizontal directions because of the limitation of our collected data. But theoretically, our network can work on any directions after modifications.}. And we will show the strong representation ability of our architecture on multi-step prediction.

\begin{figure*}[!ht]
\centering
\subfigure[Pix2Pix~\cite{pix2pix2017}]{
\begin{minipage}[b]{0.18\linewidth}
\includegraphics[width=1\linewidth]{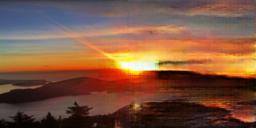}
\includegraphics[width=1\linewidth]{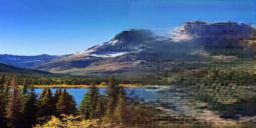}
\end{minipage}
\label{fig:cp.a}}
\subfigure[GLC~\cite{glc}]{
\begin{minipage}[b]{0.18\linewidth}
\includegraphics[width=1\linewidth]{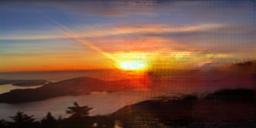}
\includegraphics[width=1\linewidth]{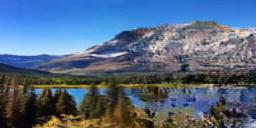}
\end{minipage}
\label{fig:cp.b}}
\subfigure[CA~\cite{generative_inpainting}]{
\begin{minipage}[b]{0.18\linewidth}
\includegraphics[width=1\linewidth]{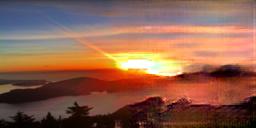}
\includegraphics[width=1\linewidth]{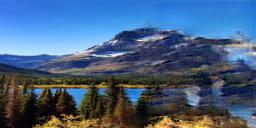}
\end{minipage}
\label{fig:cp.2018inpainting}}
\subfigure[FC+SHC]{
\begin{minipage}[b]{0.18\linewidth}
\includegraphics[width=1\linewidth]{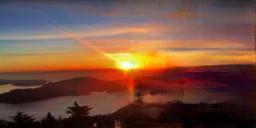}
\includegraphics[width=1\linewidth]{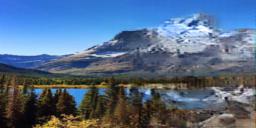}
\end{minipage}
\label{fig:cp.c}}
\subfigure[RCT+SHC (Ours)]{
\begin{minipage}[b]{0.18\linewidth}
\includegraphics[width=1\linewidth]{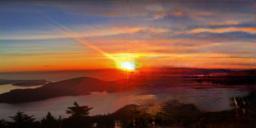}
\includegraphics[width=1\linewidth]{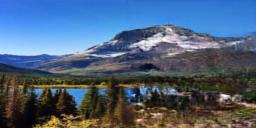}
\end{minipage}
\label{fig:cp.d}}
\caption{Comparisons on 1-step with latest generative methods. Ours RCT+SHC method achieve the best quality.}
\label{fig:cp}
\end{figure*}

\subsection{One-step Prediction}
\label{sec:4.1}
To train our model, we use Adam optimizer \cite{adam} to minimize the loss functions defined in Equation~\ref{con:7} and Equation~\ref{con:6}. We set base learning rate$=0.0001$, $\beta_1=0.5$ and $\beta_2=0.9$. Before the formal training, we set $\lambda_{adv}=0$, $\lambda_{rec}=1$ and train generator for 1000 iterations. In the formal training, we set $\lambda_{adv}=0.002$ and $\lambda_{rec}=0.998$. Same as the training method in  \cite{wgan}, the disciminator updates parameters $n_{cir}$ times but the generator once. When iterations is less than $30$ or a multiple of $500$, we set $n_{cir}=30$. In other cases, we set  $n_{cir}=5$. The batch size is $32$, and the learning rate is divided by 10 after $1,000$ epochs. The epoch number in our training process is $1,500$.

In training, each image is resized to 144$\times$432, and then a 128$\times$256 image is randomly cropped from it or its horizontal flip. 
In testing, we resize the image to 128$\times$256.

  

\begin{table}[h] 
  \centering  
    \begin{tabular}{c|c|c}  
    \hline
    \textbf{Number of GRB} & \textbf{IS} & \textbf{FID} \\ 
    \hline
    $0$ & 2.756 & 15.171   \\
    \hline
    $1$ & 2.765 & 14.828  \\
    \hline
    $3$ (ours) & \textbf{2.852} & \textbf{13.713}  \\
    \hline
    \end{tabular}
  
  \caption{Evaluation of Inception Score (IS)~\cite{is} (the higher the better) and Fr\'echet Inception Distance (FID)~\cite{fid} (the lower the better) of different number of GRB. $0$ means no GRB used in the network. $1$ means we keep the GRB where the feature size is $16\times 32 \times 256$. $3$ is the setting utilized by us.} 
  \label{table3} 
\end{table}

\textbf{Comparison with Previous Works }We make comparisons with latest generative methods\footnote{We make some modifications on their implementation for image outpainting.}, including Pix2Pix~\cite{pix2pix2017}, GLC~\cite{glc}, and Contextual Attention~\cite{generative_inpainting}, which are originally designed for image inpainting. The comparison result is shown in Fig.~\ref{fig:cp}. We can find that our method achieves the best generation quality due to our designed architecture.

\begin{figure}[!ht]
\centering
\includegraphics[width=1\linewidth]{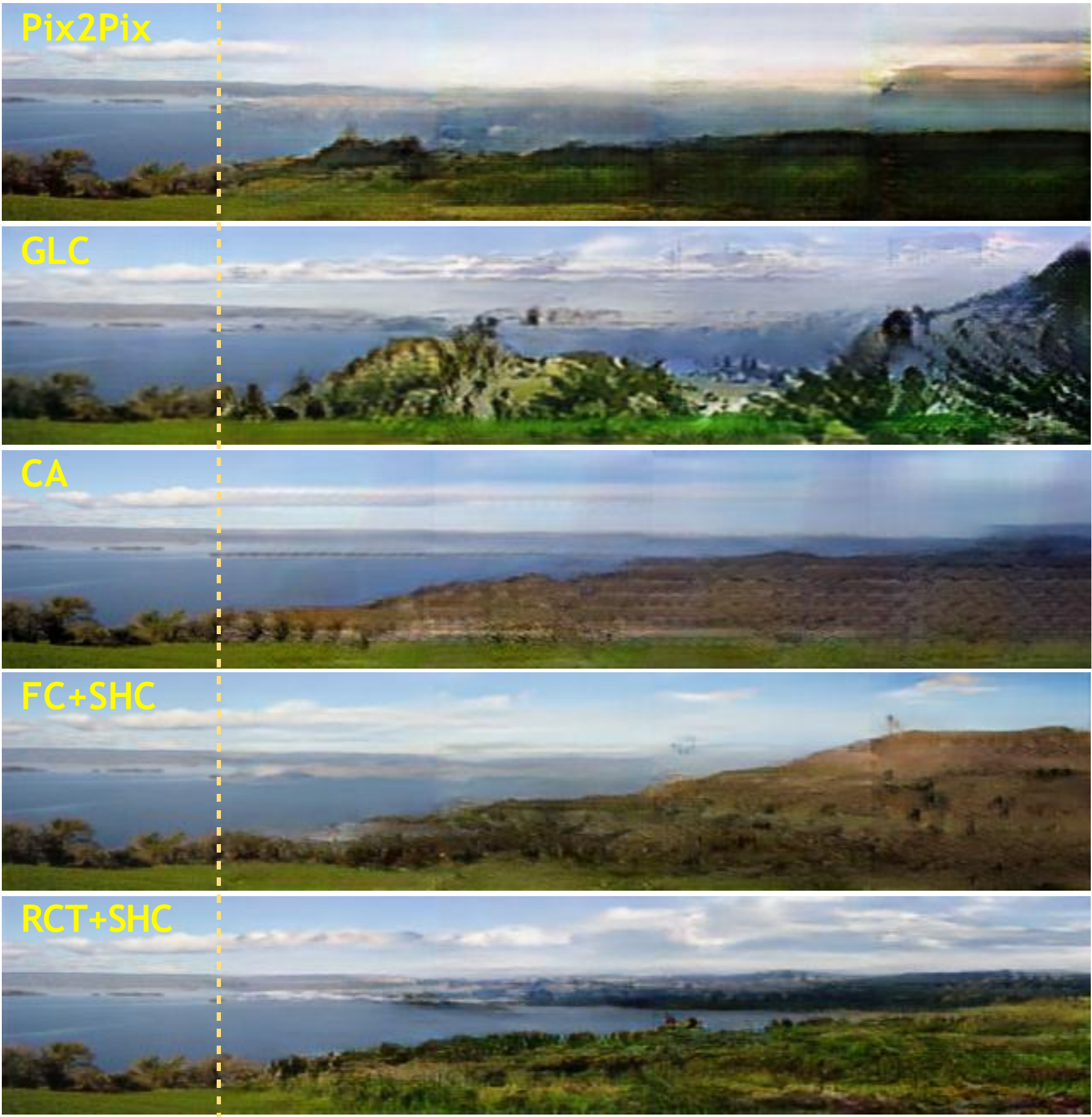}
\caption{Comparisons on multi-step predictions.}
\vspace{-3mm}
\label{fig:cp_endless}
\end{figure}

We employ Inception Score~\cite{is} and Fr\'echet Inception Distance~\cite{fid} to measure the generative quality objectively, and report them in Table.~\ref{tab:is_fid}. Our method achieves the best performance of FID, but its IS is a bit lower than CA~\cite{generative_inpainting}. This is because CA employs a contextual attention method, which uses the feature in the original region to reconstruct prediction. But as shown in Fig.~\ref{fig:cp}, ~\ref{fig:cp_endless}, the contextual attention makes predictions worse when far away from original inputs. This leads to poor FID score ($19.040$, while ours $13.713$). The contextual attention is an effective method in small region prediction (such as inpainting), but is not suitable in long-range outpainting.

\begin{table}[t!] 
  \centering

    \begin{tabular}{c|c|c}  
    \hline
    \textbf{Method} & \textbf{IS}  & \textbf{FID}  \\ 
    \hline
    Pix2Pix~\cite{pix2pix2017} & 2.825 & 19.734  \\
    \hline
    GLC~\cite{glc} & 2.812 & 14.825  \\
    \hline
    CA~\cite{generative_inpainting} & \textbf{2.931} & 19.040  \\
    \hline
    FC+SHC & 2.845 & 15.186  \\
    \hline
    RCT+SHC (Ours) & 2.852 & \textbf{13.713}  \\
    \hline
    \end{tabular}
  
  \caption{Evaluation of IS~\cite{is} and FID~\cite{fid} scores in 1-step prediction. Images from the validation set have an IS of $3.387$. We evaluate FID score between predictions and validation set which has $1,000$ images. } 
  \label{tab:is_fid}
\end{table}

\textbf{Ablation Study} First, we conduct ablation studies to demonstrate the necessity of introduction of SHC and RCT. The qualitative result comparison is shown in Fig.~\ref{fig:7}, in which we compare our architecture with the models without SHC or RCT. According to the experimental results, SHC successfully mitigates the un-smoothness between the predicted and original region. And RCT effectively improves the representation ability of the model and make the details in the prediction more delicate. Second, we make an ablation study on GRB. As shown in Table.\ref{table3}, the performance improves when using more GRB modules, which demonstrates the effectiveness.


\begin{figure*}[ht]
\centering

\includegraphics[width=0.7\linewidth]{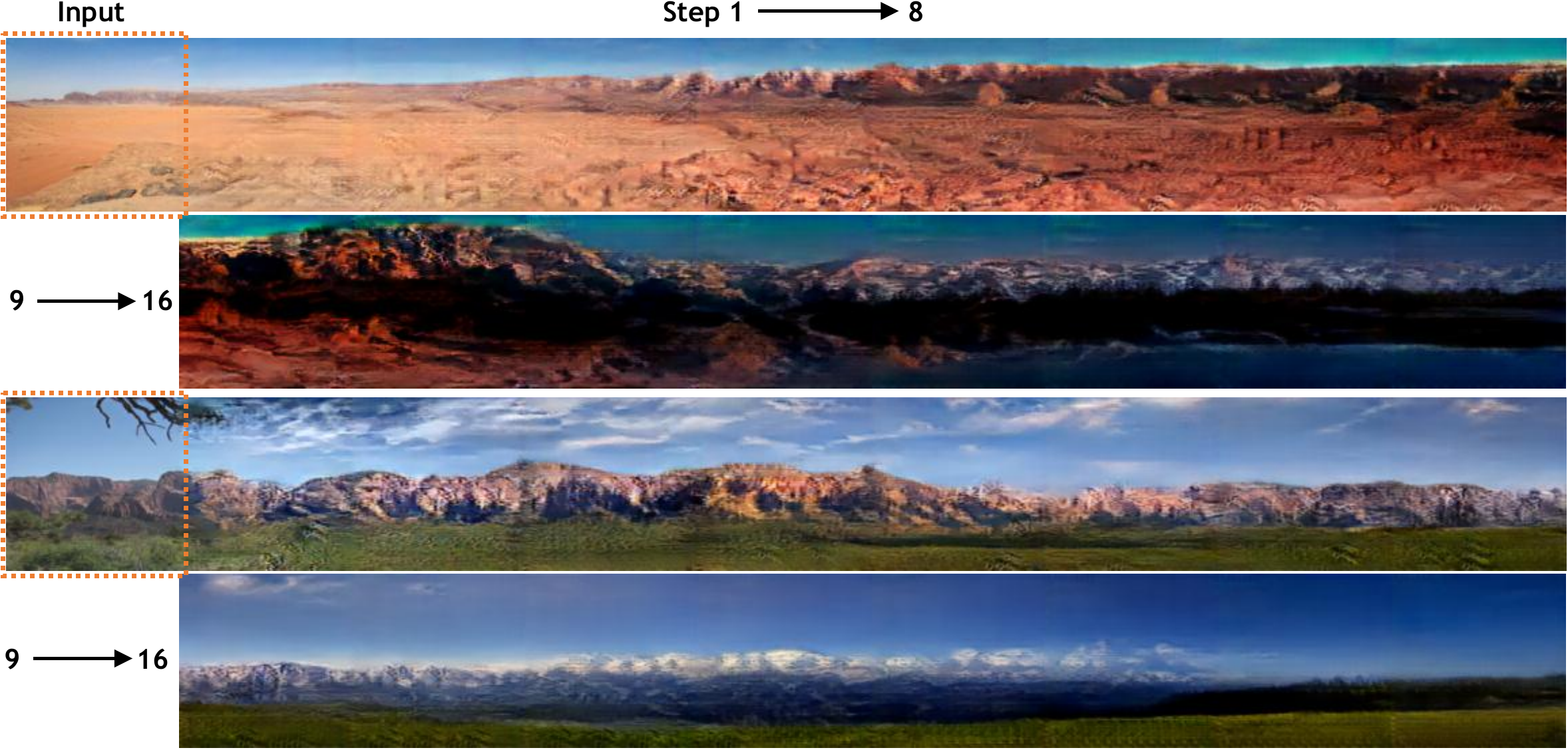}

\caption{The prediction of very long range. Given an input image (128$\times$128 size), we predict 16 steps to the right direction (128$\times$2176 size). Each example is shown in two lines.}
\vspace{-3mm}
\label{fig:9}
\end{figure*}

\begin{figure*}[!t]
\centering

\includegraphics[width=0.7\linewidth]{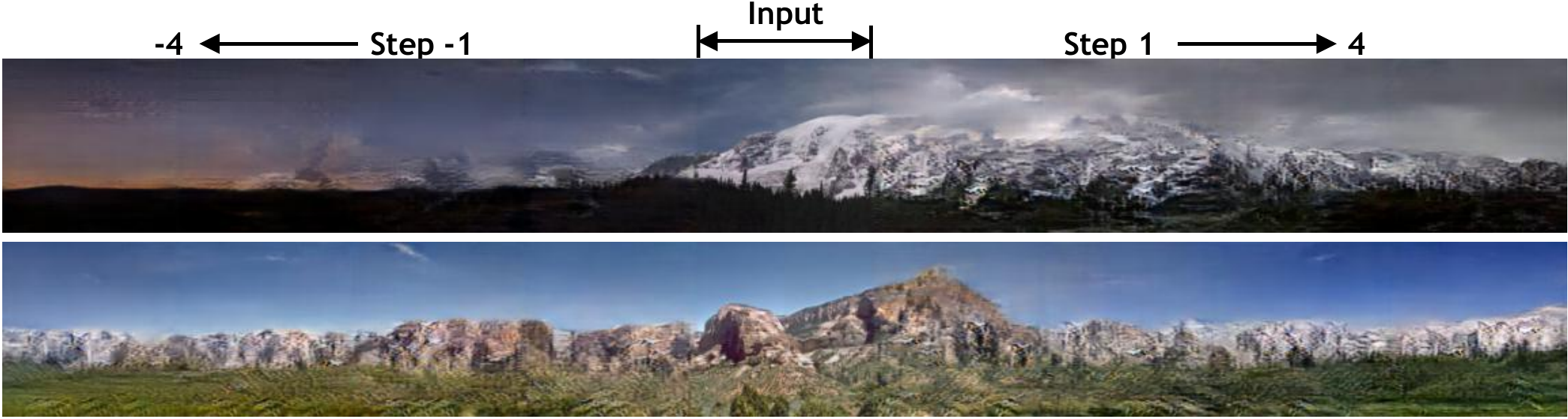}

\caption{The prediction of an input image on both sides. Given an input image (128$\times$128 size), we predict 4 steps to both the left (step: -1:-4) and right (step: 1:4) directions (128$\times$1152 size). The middle of the example is the input region.}
\vspace{-3mm}
\label{fig:10}
\end{figure*}


\subsection{Multi-Step Prediction}
In this section, we use the well-trained model in Section~\ref{sec:4.1} for multi-step prediction experiments. To make multi-step predictions, we use the predicted output from the previous step as the input for the next step. By concatenating the results from each step, we can get a very long picture.

We experiment with the prediction on one side in a very long range (Fig.~\ref{fig:9}) and the prediction on both sides (Fig.~\ref{fig:10}). These two experiments both show the powerful representational capabilities of our architecture. By the benefit of RCT, our model allows for long-term predictions with only a small amount of noise increase.

Besides, we make a comparison between our method and previous works: Pix2Pix~\cite{pix2pix2017}, GLC~\cite{glc}, and CA~\cite{generative_inpainting} on multi-step predictions. The comparison result is shown in~\ref{fig:cp_endless}. Again, the result consistency in Pix2Pix~\cite{pix2pix2017}, GLC~\cite{glc}, and CA~\cite{generative_inpainting} drops dramatically under this circumstance. FC+SHC achieves a better consistency, but still suffers from a large blurry effect. Especially, when far away from original inputs, sharp edges occur in the prediction results. By replacing the FC module with RCT, our method achieves the best performance on both consistency and sharpness.

\begin{figure}[!ht]
\centering
\vspace{-10pt}
\includegraphics[width=1\linewidth]{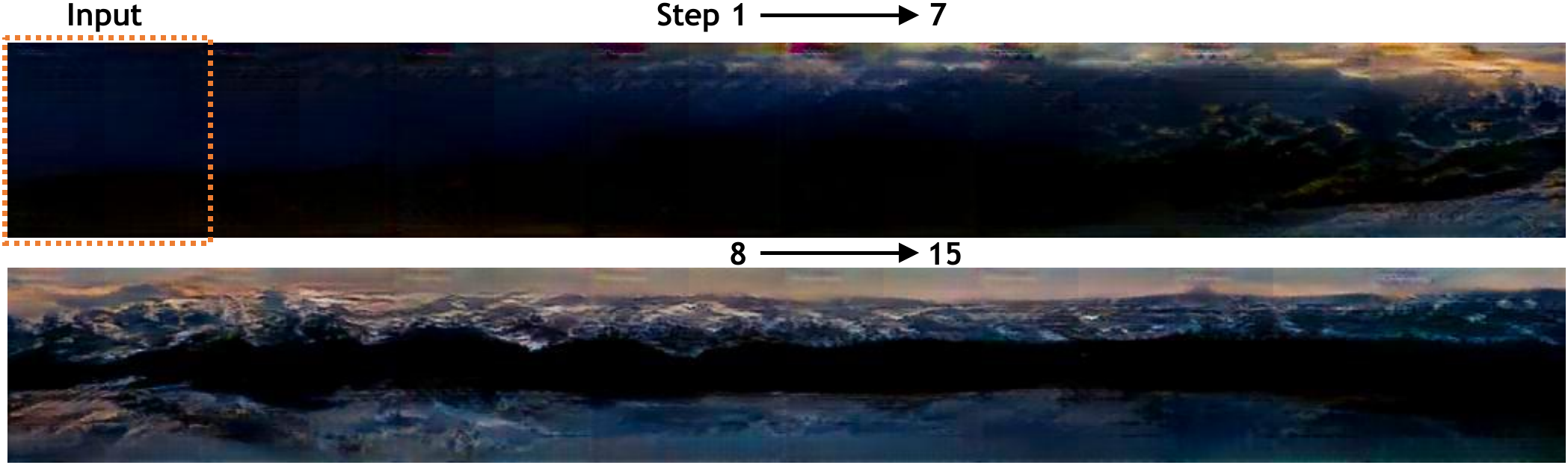}
\caption{Generation from an input with few observable details, which is a hard case for previous Non-DL methods.}
\vspace{-20pt}
\label{fig:generate_detail}
\end{figure}

\textbf{A Hard Case Example.} We test our method on some difficult cases, which are hard for previous works based on image matching. We show one example in Fig.~\ref{fig:generate_detail}. As shown in Fig.~\ref{fig:generate_detail}, when a given input is nearly nonobservable due to its darkness, our method is still able to generate a highly realistic snow mountain. 


\section{Conclusion and Future Work}
We design a novel end-to-end network to solve image outpainting problems, which is, to the best of our knowledge, the first approach to utilize a deep neural network for solving this problem. With the introduction of the graceful designed Recurrent Content Transfer, Skip Horizontal Connection, and Global Residual Block, our network can generate images with high quality and extra length. We collect a new natural scenery dataset and conduct a series of experiments on it. Not surprisingly, our proposed method achieves the best performances. More than that, the proposed method can successfully generate extremely long pictures by iterating the model, which is unprecedented. 

In future work, we would like to explore how to extrapolate images on horizontal and vertical directions with one same model simultaneously. Besides, we plan to design a specialized training process for the multi-step prediction.


{\small
\bibliographystyle{ieee_fullname}
\bibliography{egbib}
}

\end{document}